%% file: aaaimake20Survey.tex
\documentclass[letterpaper]{article}
\usepackage{aaai20}
\usepackage{times}
\usepackage{helvet}
\usepackage{courier}
\usepackage[hyphens]{url}
\usepackage{graphicx}

\usepackage{amssymb}
\usepackage{hyperref}

\usepackage{makecell}

\title{A Review on Intelligent Object Perception Methods\\Combining Knowledge-based Reasoning and Machine Learning\thanks{This project has received funding from the Hellenic Foundation for Research and Innovation (HFRI) and the General Secretariat for Research and Technology (GSRT), under grant agreement No 188.}
}

\author{Filippos Gouidis,\textsuperscript{\rm 1\thanks{Authors contributed equally.}}
Alexandros Vassiliades,\textsuperscript{\rm 1, 2$^\dagger$}
Theodore Patkos,\textsuperscript{\rm 1}\\
{\bf \Large
Antonis Argyros,\textsuperscript{\rm 1}
Nick Bassiliades,\textsuperscript{\rm 2}
Dimitris Plexousakis\textsuperscript{\rm 1}}\\
\textsuperscript{\rm 1}Institute of Computer Science, Foundation for Research and Technology, Hellas,\\
\textsuperscript{\rm 2}Aristotle University of Thessaloniki\\
\{gouidis, patkos, argyros, dp\}@ics.forth.gr,
\{valexande, nbassili\}@csd.auth.gr}

\begin{document}
\maketitle
\begin{abstract}
Object perception is a fundamental sub-field 
of Computer Vision, covering a multitude of individual areas and having contributed high-impact results. While Machine Learning has been traditionally applied to address related problems, recent studies also seek ways to integrate knowledge engineering in order to expand the level of intelligence of the visual interpretation of objects, their properties and their relations with the environment. In this paper, we attempt a systematic investigation of how knowledge-based methods  contribute to diverse object perception tasks. We review the latest achievements and identify prominent research directions.\end{abstract}


\input{intro} \label{sec:intro}
\input{representation} \label{sec:representation}
\input{commonsense} \label{sec:cs}
\input{learning} \label{sec:learning}
\input{discussion}
\label{sec:discussion}
\input{conclusions} \label{sec:conclusions}

\bibliographystyle{aaai}
\bibliography{socolaBiblio19}

\end{document}

%% file: intro.tex

\section{Introduction}







Despite the recent sweep of progress in Machine Learning (ML) which is stirring  public imagination about the capabilities of future Artificial Intelligence (AI) systems, the research community seems more composed, in part due to the realization that current achievements are largely based on engineering advancements, and only partly on novel scientific progress. Undoubtedly, the accomplishments are neither small nor temporary; in the latest ``One Hundred Year Study on AI", a panel of renowned AI experts is foreseeing tremendous impact of AI in a multitude of technological and societal domains in the next decades, fueled primarily by systems running ML algorithms~\cite{Stone16AIReport}. Yet, there is still a lot of ground to cover, before we can obtain a deep understanding of how to overcome the limitations of data-driven approaches at a more generic level.

Aiming at exploiting the full potential of AI, a growing body of research is devoted to the idea of integrating ML and knowledge-based approaches. Davies and Marcus (\citeyear{Davis15}) for instance, while discussing the multifaceted challenges related to automating commonsense reasoning, a crucial ability for any intelligent entity operating in real-world conditions, underline the need to combine the strengths of diverse AI approaches from these two fields. Others, as for example Bengio et al. (\citeyear{Bengio19}) and Pearl (\citeyear{Pearl18}), emphasize the inability of Deep Learning to effectively recognize cause and effect relations. Pearl, Geffner (\citeyear{Geffner18}) and recently Lenat\footnote{\url{https://towardsdatascience.com/statistical-learning-and-knowledge-engineering-all-the-way-down-1bb004040114}}, suggest to seek solutions by bridging the gap between model-free, data-intensive learners and knowledge-based models and by building on the synergy between heuristic level and epistemological level languages.

In this paper, we review recent progress in the direction of coupling the strengths of ML and knowledge-based methods, focusing our attention on the topic of Object Perception (OP), an important sub-field of Computer Vision (CV). Tasks related to OP are at the core of a wide spectrum of practical systems and relevant research has traditionally relied on ML to approach the related problems. The recent developments have significantly advanced the field, but, interestingly, state-of-the-art studies try to  integrate symbolic methods, in order to achieve broader visual intelligence. It seems that it is becoming less of a paradox within the CV community that in order to build intelligent vision systems, much of the information needed is not directly observable. 

Existing surveys on the intersection of ML and knowledge engineering (e.g., \cite{Nickel0TG16}) are indeed very informative, but they usually offer a high-level understanding of the challenges involved. The rich literature on CV reviews, on the other hand, adopts a more problem-specific analysis, studying in detail the requirements of each particular CV task (see e.g., \cite{wu2017visual,herath2017going,liu2018deep}). Only recently was an overview presented that shows how background knowledge can benefit tasks, such as image understanding \cite{Somak19}; our goal is to explore  this direction on the topic of intelligent OP, reporting state-of-the-art achievements and showing how different facets of knowledge-based research can contribute to addressing the rich diversity of OP tasks.



\begin{figure}[!h]
	\begin{center}
		\includegraphics[width=1.0\linewidth]{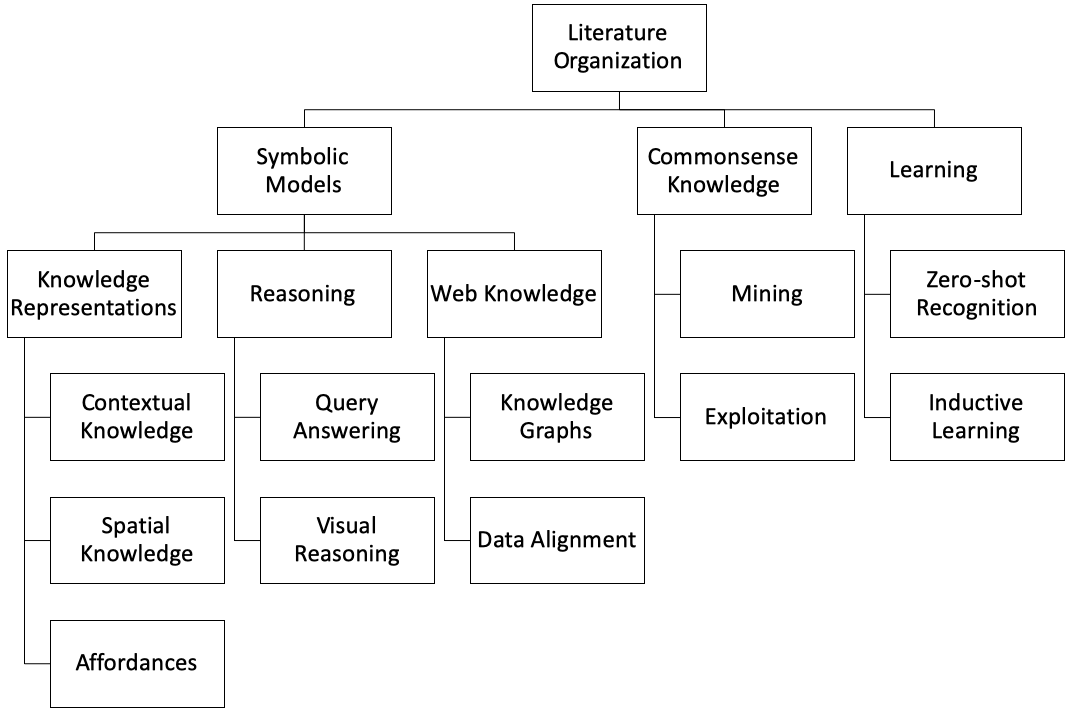}
	\end{center}
	\caption{Organization of thematic areas.}
	\label{fig:diagram}
\end{figure}

The rest of the paper investigates state-of-the-art literature on intelligent OP along the three pillars shown in Figure \ref{fig:diagram}): i) symbolic models, further analyzed from the perspectives of expressive representations, reasoning capacity, and open domain, Web-based knowledge; ii) commonsense knowledge exploitation, a key skill for any intelligent system; and, iii) enhanced learning ability, building on hybrid approaches. By no means should this review be considered exhaustive; inevitably, relevant studies may not have been included. Our intention is to capture a snapshot of the most recent research trends, discussing studies that improve previous results or offer new insights, hoping to provide a starting point for following the interesting progress currently underway, while giving pointers to directions that seem worth investigating. In this respect, the paper concludes with a discussion on open questions and prominent research directions. Table~\ref{tbl:overview} at the end summarizes the reviewed literature.

%% file: representation.tex

\section{Exploitation of Symbolic Models}


The scope of OP research ranges over a wide spectrum of problems, from object, action and affordance detection, to localization and recognition in images, to motion and structure inference in videos, to scene understanding and visual reasoning. Traditionally, OP relied on ML methodologies to find patterns in realms of data, taking as input feature vectors representing entities in terms of numeric or categorical (membership to a more general class) attributes. 

In this section, we discuss how high-level knowledge related to visual entities can improve the performance of OP algorithms in manifold ways. We start by reviewing  state-of-the-art in coupling data-driven approaches and rich knowledge representations about aspects such as context, space and affordances. 

\subsection{Expressive Representations of Knowledge}
Although the notion of a representation is rather general, the modeling of relational knowledge in the form of individuals (entities) and their associated relations is well-studied in AI, especially in the context of symbolic representations (see for instance \cite{vanHarmelenLifHandbook07,Brachman04book}). The goal is to offer the level of abstraction needed to design a system versatile enough to adapt to the requirements of a particular domain, yet rigid enough to be encoded in a computer program, having clear semantics and elegant properties \cite{Dean95book}. As a result, \textit{expressiveness}, i.e., what can or cannot be represented in a given model, and \textit{computation}, i.e., how fast conclusions are drawn or which statements can be evaluated by algorithms guaranteed to terminate, are two, often competing, aspects to be taken into consideration. 

The form of the representational model applied plays a decisive role on such considerations, affecting the richness of semantics that can be captured. While the range of models varies significantly, even relatively shallow representations have proven to offer improvements in the performance of OP methodologies (see for example \cite{deng2014large,Feifei14,zhu2015building,redmon2017yolo9000}). Models as simple as  relational tables or flat weighted graphs, but also more complex multi-relational graphs, often called Knowledge Graphs (KGs), or even semantically-rich conceptualizations with formal semantics, often called ontologies, are proposed in the relevant literature. In the sequel, we refer to any model that offers at least a basic structuring of data as a Knowledge Base (KB).

\subsubsection{Utilization of Contextual Knowledge} Context awareness is the ability of a system to understand the state of the environment, and to perceive the interplay of the entities inhabiting it. This feature offers advantages in accomplishing inference tasks, but also enhances a system in
terms of reusability, as contextual knowledge enables it to adapt to new situations and environments that resemble known ones.

A preliminary approach towards this direction utilized a special semantic fusion network, which combined novel object- and scene-based information and was able to capture relationships between video class labels and semantic entities (objects and scenes) \cite{Wu_2016_CVPR}. This was a Convolutional Neural Network (CNN) consisting of three layers, with the first layer detecting low-level features, the second layer object features and the third layer scene features. Although no symbolic representation was applied, this modeling of abstraction layers for the representation of the information depicted in an image is similar to the hierarchical structuring of knowledge used by  top-down methods.

In a similar style, Liu et al. (\citeyear{Liu2018}) recently attempted to address the problem of object detection by proposing an algorithm that exploits jointly the context of a visual scene and an object's relationships. Such features are typically taken into consideration in isolation by most object detection methods. The algorithm uses a CNN-based framework tailored for object detection and combines it with a graphical model designed for the inference of object states. A special graph is created for each image in the training set, with  nodes corresponding to objects and  edges to object relationships. The intuition behind this approach is that the graph's topology enables an object state to be determined not only by its low-level characteristics (appearance details),  but also  by the states of other objects interacting with it and by the overall scene context. 
The experimental evaluation conducted on different datasets underscored the importance of knowledge stemming from local and global context. 


Currently, a collection of prominent approaches
are oriented towards Gated Graph Neural Networks
(GGNNs) \cite{li2015gated}, a variation of Gated Neural
Networks (GNNs) \cite{scarselli2008graph}, in order to integrate contextual knowledge in the training of a system. GNNs are a special type of NNs tailored to the learning of information
encoded in a graph. In GGNNs, each node corresponds
to a hidden state vector that is updated in a iterative way. 
Two features that differentiate GGNNs from standard Recurrent Neural Networks (RNNs), which also use the mechanism of recurrence, is that in the former information can move bi-directionally and that many nodes can be updated per step.

Chuang et al. (\citeyear{Chuang2018}) propose a GGNN-based method which exploits  contextual information, such as the types of objects and their spatial relations, in order to detect action-object affordances. In the same vein, Ye et al. (\citeyear{Ye2017}) utilize a two-stage pipeline built around a CNN to detect functional areas in indoor scenes. More recently, Sawatzky et al. (\citeyear{Sawatzky2019}) utilized a  GGNN, 
which takes into account the global context of a scene and infers the affordances of the contained objects. It also proposes the most suitable object for a specific task. The authors prove that this approach yields better results compared to methods that rely solely  on the results of an object classification algorithm.

In \cite{li2017situation}, an approach aiming to perform situation recognition is presented, based on the detection of human-object interactions. The goal is to predict the most representative verb to describe what is taking place in a scene, capturing also relevant semantic information (roles), such as the actor, the source and target of the action, etc. The authors utilize a GGNN, which enables combined reasoning about verbs and their roles through the iterative propagation of messages along the edges. 

\subsubsection{Spatial Contextual Knowledge} A particular type of contextual knowledge concerns the spatial properties of the entities populating a visual scene. These may involve simple spatial relations, such as ``object $x$ is usually part of object $y$'' and ``object $x$ is usually situated near the objects $x_1,\dots,x_n$'', but also semantically enriched statements, such as ``objects of type $x$ are usually found inside object $z$ (e.g., in the fridge), located in room $y$''. Due to the ubiquitous nature of spatial data in practical domains, such relations are often captured as a separate class of context.


Semantic spatial knowledge, when fused with low-level metric information, gives great flexibility to a system. This is demonstrated in the study of Gemignani et al. (\citeyear{gemignani2016living}), where a novel representation is introduced that combines the metric information of the environment with the symbolic data that conveys meaning to the entities inhabiting it, as well as with topological graphs. Although delivering a generic model with clear semantics is not their main objective, the resulting integrated representation enables a system to perform high-level spatial reasoning, as well as to understand target locations and positions of objects in the environment.

Generality is the aim of the model proposed by Tenorth and Beetz (\citeyear{tenorth2017representations}), which uses an OWL ontology, combining information from  OpenCyc and other Web sources that help compile new classes while forming the environmental map. The main reasoning mechanisms of this study is Prolog, although probabilistic reasoners are also used to tackle fuzzy information or uncertain relations. 
A CV system annotates objects based on their shape, their distances and the dimension of the environment, 
using a monotonic Description Logic (DL), to build the environmental map. 
As a result, a coherent and well-formalized representation of environments is achieved, which can offer high-quality datasets for training data-driven models. The use of DL can also offer a wide spectrum of spatial reasoning capabilities with well-specified properties and formal semantics.

Adopting a different approach, the KG given in~\cite{Chen2018} enables spatial reasoning both in a local and a global context which, in turn, results in improved performance in semantic scene understanding. The proposed framework consists of two  distinct modules, one  focusing on local regions of the image and one dealing with the whole image. The local module is convolution-based, analyzing small regions in the image, whereas the basic component of the global module is a KG representing regions and classes as nodes, capturing spatial and semantic relations. 
According to the experimental evaluation performed on the ADE and Visual Genome datasets, the network achieves better performance over other CNN-based baselines for region classification, by a margin which sometimes is  close to 10\%. According to the ablation study, the most decisive factor for the framework's performance was the KG.

\subsubsection{Modeling Affordances} 
Building on the geometrical structure and physical properties of objects, such as rigidity and hollowness, the representation of affordances helps develop systems that can reason about how human-level tasks are performed. While ML is invaluable for automating the process of learning from example when  data is available, rich representations can generalize and reuse the obtained models in situations where data-based training is not possible. 

One of the first studies that demonstrated that even very basic semantic models can improve  the performance of recognizing  human-object interaction was~\cite{Chao2015}. The authors succeeded in boosting the performance of visual classifiers by exploiting the compositionality and concurrency of semantic concepts contained in images. 

KNOWROB 2.0~\cite{beetz2018know}, which is the result of a series of research activities in the field of Cognitive Robotics, is an excellent example of integrating top-down knowledge engineering with bottom-up information structuring, involving, among others, a variety of CV tasks. A combination of KBs of different granularity helps the KNOWROB 2.0 framework capture rich models of the world. 
 The representation of high-level knowledge is based on OWL-DL ontologies, a decidable fragment of First-order Logic (FOL), yet adequately expressive for most practical domains. The KBs enable the system to answer questions, such as \emph{``how to pick up the cup'', ``Which body part to use''} etc. The authors provide evidence that learning human manipulation tasks on existing methods can be boosted by using symbolic level structured knowledge. 


A recently proposed novel representation model that manages to balance between concept abstraction, uncertainty modeling and scalability is given in~\cite{daruna2019robocse}. The so called RoboCSE framework encodes the abstract, semantic knowledge of an environment, i.e., the main concepts and their relations, such as location, material and affordance, obtained by observations, simulations, or even from external sources, into \textit{multi-relational embeddings}. These embeddings are used to represent the knowledge graph of the domain in vector space, encoding vertices that represent entities as vectors and edges that represent relations as mappings. While the majority of similar approaches rely on Bayesian Logic Networks and Markov Logic Networks, suffering from well-known intractability problems, the authors prove that their model is highly scalable, robust to uncertainty, and generalizes learned semantics.

Learning from demonstration, or imitation learning, is a relevant, yet broader objective, which introduces interesting opportunities and challenges to a CV system (see \cite{Torabi19,ravichandar2019lfd}). Purely ML-based methods constitute the predominant research direction, and only few state-of-the-art studies utilize knowledge-based methods, taking advantage of the reusability and generalization of the learned information. A popular choice is to deploy expressive OWL-DL \cite{ramirez2017transferring,lemaignan2017artificial} or pure DL \cite{agostini2017efficient} representations to capture world knowledge. The CV modules are assigned the task to extract information about the state of the environment, the expert agent's pose and location, grasping areas of objects, affordances, shapes etc. On top of these, the coupling with knowledge-based systems assists in visual interpretation, for example to track human motion, to semantically annotate  the movement (i.e., ``how the human performs the action'') or to understand if a task is doable in a given setting. These studies show that such representations enable a system to reuse the learned knowledge in diverse settings and under different conditions, without having to re-train classifiers from scratch. Moreover, complex queries can be answered, a topic discussed in the next subsection.

\subsection{Reasoning over Expressive KBs}
Encoding knowledge in a semantically structured way is only part of the story; 
a rich representation model can also offer inference capabilities to a CV system, which are needed for accomplishing complex tasks, such as scene understanding, or simpler tasks under realistic conditions, such as scene analysis with occlusions, noisy or erroneous input etc. A reasoning system can be used to connect the dots that relate concepts together when only partial observation is available, especially in data-scarce situations, where annotated data are not sufficiently many. In such situations, the compositionality of information, an inherent characteristic of the entities encountered in visual domains, can be exploited by applying reasoning mechanisms.

\subsubsection{Complex Query Answering}
Probably the field that highlights more clearly the needs and challenges faced by a CV system in answering complex queries about a visual scene is the field of Visual Question Answering (VQA). VQA was recently introduced as a collection of benchmark image-based open-domain questions that, in order to be answered, call for a deep understanding of the visual setting. VQA goes beyond traditional CV, since apart from   image analysis, the proposed methods apply also a repertoire of AI techniques, such as Natural Language Processing, in order to correctly analyze the textual form of the question, and inferencing, in order to interpret the purpose and intentions of the entities acting in the scene~\cite{Krishna2017}. The challenges posed by this field are complex and multifaceted, a fact which is also demonstrated by the rather poor performance of state-of-the-art-systems in comparison to humans. VQA is probably the area of CV that has drawn the most inspiration from symbolic AI approaches to date. 

An indicative example is the approach recently presented by Wu et al. (\citeyear{Wu2018}), who introduced a VQA model combining observations obtained from the image with information extracted from a general KB, namely DBpedia. Given an image-question
pair, a CNN is utilized to predict a set of attributes from the image, i.e., the most recognizable objects in the image, in terms of clarity and size. Consequently,  a series of captions based on the attributes is generated, which is then used to extract relevant information from DBpedia through appropriately formulated queries. In a similar style, in \cite{narasimhan2018straight} an external RDF repository is used to retrieve properties of visual concepts, such as category, used for, created by, etc. The technique utilizes a Graph Convolution Network
(GCN), a variation of GNN, before producing an answer. In both cases, the ablation analysis reveals the impact of the KB in improving performance.

Other types of questions in VQA require inferencing about the properties of the objects depicted in an image. For example, queries such as \emph{``How is the man going to work?"} or more complex queries, such as \emph{``When did the plane land?"}, have been the subject of the study presented by Krishna et al. (\citeyear{Krishna2017}), who introduced the Visual Genome dataset and a VQA method. In fact, this is one of the first studies to bring a model trained on an RDF-based scene graph that had good recall results to all \emph{What, Where, When, Who, Why, How} queries. Even further, Su et al. (\citeyear{Su2018}) introduced the visual knowledge memory
network (VKMN) in order to handle questions, whose answers
 cannot be directly inferred from the image
visual content but require reasoning over structured human knowledge.

The importance of capturing the semantic knowledge in VQA collections led also to the creation of the Relation-VQA dataset \cite{Lu2018}, which extends Visual Genome with a special module measuring the semantic similarity of images. In contrast to methods mining  only  concepts or attributes, this model extracts  relation facts related to both concepts and attributes. The experimental evaluation conducted on VQA and COCO dataset showed that the method outperformed other state-of-the-art ones. Moreover, the ablation studies show that the incorporated semantic knowledge was crucial for the performance of the network.

Despite its increasing popularity, the VQA field is still hard to confront. The generality of existing methods is also questioned~\cite{Goyal2019}. 
Developing generic solutions, less tightly coupled to specific datasets, will definitely benefit the pursuit towards broader visual intelligence.  

 \subsubsection{Visual Reasoning} A task related to  VQA  that has gained popularity in recent  years is that of Visual  Reasoning 
  (VR). In this case, the type of questions that have to be answered are more complex and require a multi-step reasoning procedure. For example, given an image containing objects of different shapes and color, the task of recognizing the color of an object of certain shape that lies in a certain area w.r.t. the position of another object of  certain shape and color falls to the category of VR (in this case,  first  the ``source'' object must be detected, then  the ``target'' object, and, finally, its color must be recognized).
Similar to the case of VQA, a number of VR works has drawn inspiration from symbolic AI-based ideas.

In general, many VR works are based on Neural Module Networks (NMNs) which are NNs of adaptable architecture, the topology of which is determined by the parsing of the question that has to be answered. NMNs simplify  complex questions into simpler sub-questions (sub-tasks), which can be more easily addressed. The modules that constitute the MNMs are pre-defined neural networks that implement the  functions that are required for the tackling of sub-tasks, which are assembled into a layout dynamically.  Central to many MNMs  is the utilization of prior \textit{symbolic} (structured) knowledge, which  facilitates  the handling of the sub-tasks.

Hu et al. (\citeyear{Hu2017}) propose End-to-End Module Networks as a variation of NMNs. The network first uses  coarse functional expressions describing the structure of the  computation required for the answering and, then, refines it according to the textual input in order to assemble the network. For example, for the  question \textit{``how many other objects of the same size as the purple cube exist?}'', first   crude functional expression for counting and relocating would be predicted as relevant to the answering of the question which, subsequently,  would be refined by  the parameters  from text analysis (in this case one such parameter is the color of the cube).

Similarly, Johnson et al. (\citeyear{johnson2017inferring}) propose  a variation of  NMNs, which  is  based on  the concept of~\textit{programs}. Programs are symbolic structures of certain specification written in a Domain-Specific Language and are defined by a syntax and  semantics. In the context of VR, programs  describe   a sequence of functions that must be executed, in order for an answer to be computed. During testing on the CLEVR dataset the model exhibited notable performance,  generalizing better in a variety of settings, such as for new question types and human-posed questions. Building on the notion of programs, Yi et al. (\citeyear{Yi2018}) further incorporated knowledge regarding the structural scene representation of the image. The method achieved near-perfect accuracy, while also providing transparency to the reasoning process.
 

An alternative NN-based approach for  VR is found in~\cite{Santoro2017}, where the incorporation of Relation Networks (RNs) in  CNNs and Long Sort-Term Memory (LSTM) architectures is proposed. RNs are architectures whose computations focus explicitly on relational reasoning and are characterized by three important features: they can infer relations, they are data efficient, and they operate on a set of objects, a  flexible symbolic input format that is agnostic to the kind of inputs it receives. For example, an object could correspond to the background, to a particular physical object, a texture, conjunctions of physical objects etc. 

To conclude, it is worth indicating also a recent trend in visual explanation approaches that couples data-driven systems with Answer Set Programming (ASP). ASP is a non-monotonic logical formalism oriented towards hard search problems. A number of studies have emerged that combine ASP abductive or inductive reasoning for the VQA domain, especially for cases when training data are not many (see e.g., \cite{SuchanMehul17,RileySridharan19,BasuShakerinGupta20})

\subsection{The Web as a Problem-Agnostic Source of Data}
As the recent renaissance in AI is partly due to the availability of big volumes of training data, along with the computational power to analyze them, it is only reasonable to expect that data-driven approaches will turn their attention to the Web in order to collect the data needed. Although the benefits mentioned in the previous sections are still achievable, the challenges faced when using a Web repository rather than a custom-made KB are now different.

The vast majority of large-scale Web repositories are not problem-specific, containing a lot of irrelevant information for a ML system to be trained correctly. For the time being, ML systems are highly specific, excelling only when trained for a particular task and tested on similar to the training conditions. As a result, state-of-the-art approaches try to rely on the semantics of structured KBs, in order to filter out noisy or irrelevant knowledge, by integrating external knowledge when visual information is not sufficiently reliable for conclusion making.

\subsubsection{Exploitation of Web-based Knowledge Graphs and Semantic Repositories}

There exists a multitude of studies that use external knowledge from structured or semi-structured Web resources, in order to answer visual queries or to perform cognitive tasks. A characteristic example is found in \cite{li2017incorporating}, where the ConceptNet KG, a semantic repository of commonsense Linked Open Data, is used to answer open domain questions on entities 
such as ``\textit{What is the dog's favorite food?}''. The approach proceeds in a step-wise manner: first, visual objects and keywords are extracted from an image, using a Fast-RCNN for the objects and a LSTM for the syntactical analysis; then, queries to ConceptNet provide properties and values for the entities found in the image. When an answer is considered correct, a Dynamic Memory Network, which is an embedding vector space that contains vector representations of symbolic knowledge triples, is renewed for future encounter of the same query. In a rather similar style, Wu et al. (\citeyear{wu2016ask}) extract properties from DBpedia, by retrieving and performing semantic analysis on the comment boxes of relevant Wikipedia pages. Here, a CNN performs object detection on the image, whereas a pre-trained RNN correlates attributes to sentence descriptions. 

The approach presented in~\cite{shah2019kvqa} is the first attempt to answer a more knowledge-intensive category of questions, such as \textit{``Who is to the left of Barack Obama?}'' or `\textit{`Do all the people in the image have a common occupation?''}. These questions make reference to the named entities contained in an image, e.g., Barack Obama, White House, France etc. and require large KBs to retrieve the relevant information. In this case, the authors choose Wikidata, an RDF repository. They first extract named entities and then try to connect them with a Wikidata entity using SPARQL queries. In addition, they extract  spatial relations with other entities shown in the image and feed them to a Bi-LSTM. A multi-layered perceptron calculates the prediction for an answer, taking as input the output of the LSTM, along with the SPARQL results.

\subsubsection{Aligning Data Obtained from Diverse Online Sources}

Entity resolution, also known as instance matching, concerns the task of identifying which entities across different KBs refer to the same individual. As the Web is growing in size, this problem is becoming crucial, especially in application domains that need to integrate and align knowledge obtained from various sources. An increasing number of CV studies face this problem, in an attempt to interpret visual information based on commonsense, non-visual knowledge.

Two characteristic approaches are given in \cite{chernova2017situated} and \cite{young2017making} that try to assign labels to a visual scene using Bayesian Logic Networks (BLNs) and relying on commonsense knowledge.
In \cite{chernova2017situated}, knowledge is extracted from WordNet, ConceptNet, and Wikipedia. WordNet is utilized in order to disambiguate seed words returned by the CV annotator with the aid of their hypernym. ConceptNet properties, such as $IsLocatedIn$ or $UsedFor$ that may point the location of an object, are also retrieved. With this method, the system can generate a compact semantic KB given only a small number of objects. 

In \cite{young2017making}, a CNN trained on ImageNet is used to annotate objects recognized in images. The system is capable of assigning semantic categories to specific regions, by relying on DBpedia comment boxes to calculate the semantic relatedness between objects. As expected, high accuracy of such an approach is difficult to achieve, due to the diversity of information retrieved from DBpedia; consequently, smarter ways of identifying only the relevant part of the comment boxes need to be devised.

%% file: commonsense.tex

\section{Exploitation of Commonsense Knowledge}

Much of the information presented in a visual scene is not explicitly related with the features captured at the pixel level, but concerns observations implicitly depicted in images. Understanding the structure and dynamics of visual entities requires being able to interpret the semantic and commonsense (CS) features that are relevant, in addition to the low-level information obtained by photorealistic rendering techniques~\cite{vedantam2015learning}. This is a popular conclusion reached within the CV community in the pursue towards achieving visual intelligence. There is a long line of studies that attempt to address the problem of extracting commonsense knowledge from visual scenes or, similarly, of utilizing commonsense inferences to improve scene understanding. In this section, we discuss state-of-the-art approaches that advance the field in these two directions.

\subsection{Mining Commonsense Knowledge from Images}

Even though ML is becoming part of many systems, it is still not able to easily capture CS knowledge from the perceived information. Additional techniques need to be devised to extract this valuable type of knowledge from visual scenes. A combination of textual and visual analysis, which extracts \emph{subject-predicate-object triples (SPO)} about objects recognized in a scene, is addressed in certain studies, e.g., \cite{vedantam2015learning,lin2015don}.  ML classifiers for object recognition are trained on image datasets, while pre-trained NN classifiers help extract SPO triples, by considering both the entities identified by the classifiers and the textual description of the images.

In a different direction, in \cite{sadeghi2015viske} the authors rely on Web images to verify the validity of simple phrases, such as \textit{``horses eat hay''}, analyzing the spatial consistency of the relative configurations of the entities and the relations involved. This unsupervised method is particularly interesting, due to the leverage it offers in automatically enriching CS repositories. In fact, the authors show how CV-based analysis can help improve recall in KBs, such as WordNet, Cyc and ConceptNet, offering a complementary and orthogonal source of evidence.

Aditya et al. (\citeyear{Aditya2018}) 
 address the problem of generating  linguistic descriptions of images  by utilizing a special type of graph, namely scene description graphs (SDGs). Such graphs are built by  using both low-level information derived using perception methods and high-level features capturing CS knowledge stemming from  the image annotations and  lexical ontological knowledge from Web resources. SDGs produce  object, scene and constituent detection tuples, accompanied by a confidence score; pre-processed background knowledge helps remove noise contained in the detection. A Bayesian Network is utilized, in order for  the dependencies among co-occurring entities and knowledge regarding abstract visual concepts to be captured. Experimental evaluations of the method on the image-sentence alignment quality, i.e., how close the generated description is to the image being described, on Flickr8k, 30k and COCO datasets, showed that the method achieves comparable performance to previous state-of-the-art methods.

\subsection{Commonsense Knowledge in Addressing OP Tasks}


State-of-the-art CS-based methodologies improve the performance of a CV system, mainly by taking into account textual descriptions about the entities found in a visual scene or by retrieving semantic information from external sources that is relevant to the image and the task at hand. 

A combination of external Web-based knowledge, text processing and vision analysis is at the core of the study presented in \cite{wang2018fvqa}. The framework annotates objects with a Fast-RNN, trained over the MS COCO dataset. The extracted entities are enriched with (i) knowledge retrieved from Wikipedia, in oder to perform entity classification; (ii) knowledge from WebChild, attempting a comparative analysis between relevant entities; and (iii) CS knowledge obtained from ConceptNet, to create a semantically rich description. The enriched entity is stored in an RDF graph and is used to address a variety of tasks. For instance, the framework has achieved improved accuracy in VQA benchmarks, but also it can be used to generate explanations for its answers. Prominent recent studies, as in~\cite{li2019visual} and \cite{narasimhan2018straight}, also build on the direction of combining textual and visual analysis with the help of knowledge obtained from CS repositories.

Another problem that researchers try to address with the help of CS knowledge is the sparsity of categorical variables in the training datasets. For example, Ramanathan et al. (\citeyear{Ramanathan2015}) utilize a neural network framework that uses different types of cues (linguistic, visual and logical)  in the context of  human actions identification. 
Similarly, Lu et al. (\citeyear{Lu2016}) exploit
language priors extracted from the semantic features of an image, in order to facilitate   the  understanding of visual relationships. The proposed model combines  a visual module tailored
to the learning of visual appearance models for  objects and predicates with a language module capable of detecting  semantically  related relationships.
 
More recently, Gu et al. (\citeyear{gu2019scene}) utilize  commonsense knowledge stemming from an  external KB in the context of scene graph generation. Namely, a special knowledge-based feature refinement module is used, which incorporates CS knowledge  from ConceptNet for the prediction of object labels consisting of triplets containing the top-K corresponding relationships, the object entity and a weight corresponding to the frequency of the triplet. This strategy, aiming to address the long tail distribution of relationships, differentiates the approach from the linguistic-based ones described previously, managing to showcase improvement in generalizability and accuracy. 
 
CS knowledge is also used to tackle other CV problems, such as in understanding relevant information about unknown objects existing in a visual scene. In \cite{icarte2017general} or \cite{young2016towards} for instance, external CS Web-based repositories are used as a source for locating relevant information. The general idea in both approaches is to retrieve as much information as possible about the recognizable objects that, based on diverse metrics, are considered semantically close to the unknown ones. $RelatedTo$, $IsA$, $UsedFor$ properties found in ConceptNet, or comment boxes retrieved from DBpedia are all relevant knowledge that can be used for developing semantic similarity measures. Similar to some extent, is the approach presented in \cite{ruiz2016probability}, which relies on RDF graphs with a probabilistic distribution over relations to capture the CS knowledge, but reverts also to a human-supervised learning approach whenever unknown objects are encountered.

%% file: learning.tex

\section{Ability to Learn New Knowledge}

The majority of state-of-the-art studies covered in the previous sections exploit a loosely-coupled combination of ML and knowledge-based methodologies. A tighter integration of methodologies of the two fields is expected to achieve much broader impact, especially in the process of learning. In the sequel, we consider  prominent attempts towards this direction, originating  either from a model-free standpoint or from a more declarative, inductive-based perspective.

\subsection{Model-Free Learning}


Recent studies devise methods that attempt to exploit information contained in higher-level representations, in order to improve scalability and generalization for tasks, such as Zero-Shot Learning (ZSL). ZSL is the problem of recognizing objects for which no visual examples have been obtained and is typically achieved by exploring a semantic embedding space, e.g., attribute or semantic word vector space.


For example, Fu et al. (\citeyear{fu2015zero})  utilize a semantic class label graph, which results in a more accurate distance metric in the semantic embedding space and an improved performance in ZSL. 
Likewise, Xian et al. (\citeyear{xian2016latent}) address the same problem by proposing a novel latent embedding model, which  learns a compatibility function between the image and semantic (class) embeddings. The model utilizes image and class-level side-information that is either collected through human annotation or through an unsupervised way from a Web repository of text corpora.

Lee et al. (\citeyear{Lee2018}) propose a novel deep learning architecture for multi-label ZSL, which relies on KGs for the discovery of the  relationships between multiple classes of objects.  The KG is built on knowledge stemming from   WordNet  and contains
3 types of label relations, super-subordinate, positive correlation, and negative correlation. The KG is coupled to a GGNN-type module for predicting labels. 

In the same vein, Wang, Ye and Gupta (\citeyear{Wang2018b}) exploit the information contained in KGs about unseen objects, in order to infer visual attributes that enable their detection.  The KG nodes correspond  to  semantic categories and the edges to semantic relationships, whereas the input to each node is the vector representation (semantic embedding)
of each category. A GCN is used to transfer information  between different categories. 
This way, by utilizing the   semantic embeddings of a novel category, the method can link categories in the KG to familiar ones  and, thus, infer its attributes. The experimental evaluation demonstrated a significant improvement on the ImageNet dataset, while the ablation studies indicated that  the incorporation of KGs enabled the system to learn meaningful classifiers on top of semantic embeddings.

In \cite{Marino2017},  the use of structured prior knowledge led to  improved performance on the task of  multi-label image classification.  The KG is built using WordNet for the concepts and Visual Genome for the relations among them. An interesting aspect of this study is the introduction of a novel NN architecture, Graph Search Neural Network, as a means to efficiently incorporate large knowledge graphs, in order to be exploited for CV tasks. 

\subsection{Inductive Learning}
The benefits of developing intelligent visual components with reasoning and learning abilities are becoming evident in broader to CV domains, such as in the field of Robotics. This conclusion was nicely demonstrated in a recent special issue of the AI Journal~\cite{AIRoboticsSP17}, where causality-based reasoning emerged as a key contribution. It is, therefore, interesting to investigate how the recent trend in combining knowledge-based representations with model-free models for the development of intelligent robots is making an impact in related OP research.


A highly prominent line of research for modeling uncertainty and high-level action knowledge is focusing on combining expressive logical probabilistic formalisms, ontological models and ML. In~\cite{Antanas18} for example, the system learns probabilistic first-order rules describing relational affordances and pre-grasp configurations from uncertain video data. It uses the ProbFOIL+ rule learner, along with a simple ontology capturing object categories. 

More recently, Moldovan et al. (\citeyear{Moldovan2018}) significantly extended this approach, using the Distributional Clauses (DCs) formalism that integrates logic programming and probability theory. DCs can use both continuous and discrete variables, which is highly appropriate for modeling uncertainty, in comparison for instance to ProbLog, which is commonly found in relevant literature. Compared to approaches that model affordances with Bayesian Networks, this approach scales much better, but most importantly, due to its relational nature, structural parts of the theory, such as the abstract action-effect rules, can be transferred to similar domains without the need to be learned again.

A similar objective is pursued by Katzouris et al. (\citeyear{Katzouris19}), who propose an abductive-inductive incremental algorithm for learning and revising causal rules, in the form of Event Calculus programs. The Event Calculus is a highly expressive, non-monotonic formalism for capturing causal and temporal relations in dynamic domains. The approach uses the XHAIL system as a basis, but sacrifices completeness due to its incremental nature. Yet, it is able to learn weighted causal temporal rules, in the form of Markov Logic Networks, scaling up to large volumes of sequential data with a time-like structure.

Also worth mentioning is the study of Antanas et al. (\citeyear{antanas2018semantic}), which instead of learning how to map visual perceptions to task-dependent grasps, it uses a probabilistic logic module to semantically reason about the most likely object part to be grasped, given the object properties and task constraints. The approach models rules in Causal Probabilistic logic, implemented in ProbLog, in order to reason about object categories, about the most affordable tasks and about the best semantic pre-grasps. 

%% file: discussion.tex

\section{Open Problems and Research Questions}

\begin{table*}
	\caption{Overview of the reviewed literature}
	\vspace*{0.5cm}
	\label{tbl:overview}
	\centering
	\small
	\begin{tabular}{|p{5cm}|p{2cm}|p{2.2cm}|p{2cm}|p{1.3cm}|p{2.5cm}| }
		\hline
		\textbf{Indicative Recent Literature} & \textbf{CV Problem Focus}  &  \textbf{ML Methods Applied} & \textbf{KB Methods Applied} & \textbf{KB Contribution} & \textbf{KB-ML Impact}\\
		\hline
		\hline

		\cite{Chuang2018}, \cite{Ye2017}, \cite{Sawatzky2019}, \cite{Chao2015}, \cite{Ramanathan2015} & affordance detection & CNN, GNN, GGNN  & Knowledge Graphs & 3, 4, 5, 7 & offers new insights\\
		\cite{beetz2018know}, \cite{ramirez2017transferring}, \cite{lemaignan2017artificial}, \cite{agostini2017efficient}, \cite{Moldovan2018} & affordance detection & scoring functions, probabilistic programming models, Bayesian Networks & OWL Ontology & 1, 2, 3, 4, 5, 6, 9 & offers new insights and improves SotA\\
		
		\hline
		\cite{icarte2017general}, \cite{redmon2017yolo9000}, \cite{Liu2018} & object detection & RCNN, CNN & Knowledge Graph, BLN & 1, 3, 4, 5, 8 & offers new insights\\
		\cite{gemignani2016living}, \cite{tenorth2017representations}, \cite{young2016towards}, \cite{beetz2018know}   & object detection & scoring functions, probabilistic programming models & OWL Ontology, DL, MLN & 1, 2, 3, 4, 5, 8, 9 & improves SotA\\
		
		\hline
		\cite{chernova2017situated}, \cite{young2017making}, \cite{Aditya2018} 
		& scene understanding & probabilistic programming, Bayesian Network & BLN & 2, 3, 4, 8 & offers new insights\\
		\cite{gu2019scene}, \cite{li2017situation}, \cite{Chen2018} & scene understanding & GGNN & Knowledge Graph & 3, 4, 7 & improves SotA\\

		\hline
		\cite{Krishna2017}, \cite{Zhu2015a}, \cite{li2019visual}, \cite{wu2016ask}, \cite{Wu2018}, \cite{li2017incorporating}, \cite{sadeghi2015viske}, \cite{shah2019kvqa}, \cite{Su2018}, \cite{narasimhan2018straight}, \cite{wang2018fvqa}  & VQA & CNN, LSTM, RCNN & Knowledge Graphs (RDF mostly) & 1, 2, 3, 4, 5, 8 & offers new insights and improves SotA \\
		 \cite{vedantam2015learning}, \cite{lin2015don} & VQA & Gausian Mixture Model, SVM &  RDF Graph & 2, 3, 4, 5  & improves SotA\\
		 \cite{Lu2018} & VQA & Gated Recurrent Unit Network &  RDF Graph & 1, 2, 4, 8  & offers new insights\\

		\hline
		\cite{Hu2017}, \cite{johnson2017inferring}, \cite{Yi2018}, \cite{Santoro2017} & visual reasoning & Neural Module Network & Symbolic Programming Language & 2, 3, 5 & offers new insights\\
		
		\cite{SuchanMehul17}, \cite{RileySridharan19}, \cite{BasuShakerinGupta20} & visual reasoning, VQA & CNN, RCNN & Non-monotonic logics, ASP & 2, 3, 4, 5, 6, 7, 9 & offers new insights\\
		
		\hline
		\cite{Marino2017}, \cite{Lee2018}, \cite{Wang2018b}  & image classification/ zero-shot recognitions & GGNN, GCN,  & Knowledge Graph, RDF Graph & 1, 2, 5 &  offers new insights and improves SotA\\
		
		\cite{fu2015zero}, \cite{xian2016latent}  & image classification/ zero-shot recognitions & Latent embedding model, Markov Chain Process & Knowledge Graph & 1, 2, 5 &  offers new insights\\

		\hline
	\cite{Antanas18}, \cite{antanas2018semantic}, \cite{Moldovan2018}, \cite{Katzouris19}   & affordance learning & scoring functions, probabilistic programming models & FOL, Causal Probabilistic Logic, MLN, Event Calculus & 1, 2, 3, 6, 7, 9 & improves SotA\\

		\hline\hline

		\multicolumn{6}{l}{\textbf{KB Contribution}: 1:concept abstraction/reuse, 2:complex data querying, 3:spatial reasoning, 4:contextual reasoning,  }\\
		\multicolumn{6}{l}{5:relational reasoning, 6:temporal reasoning, 7:causal reasoning, 8:access to open-domain knowledge, 9:formal semantics}\\
	\end{tabular}
\end{table*}

The review of the state-of-the-art reveals prominent solutions for various OP-related topics, as well as novel contributions that offer new insights (Table~\ref{tbl:overview}). The analysis can also help frame open questions towards combining ML and knowledge-based approaches in the given context. 

\subsection{Obtaining Human Commonsense}

The exploitation of CS knowledge is a characteristic example of a still open research area. Its significance was acknowledged more than two decades ago and the research conducted over the years contributed methods that combine the strengths from diverse fields of AI. At the same time, it is evident that there is still a long way to go; just the coupling of textual and visual embeddings, the mainstream in current VQA related studies, has proven to be a challenging task. Further directions need to also be explored, such as in performing complex forms of CS inferencing or in fusing the huge volume of general knowledge that exists on the Web, while eliminating the bias of information found online.  

Progress in the field of learning from demonstration can prove a vital contribution to CS inferencing and vice versa. Leaving the visual challenges involved aside, this application domain, characterized by the central role of, human mostly, agents, offers theory building opportunities on diverse perspectives. Interaction with human users calls for intuitive means of communication, where high-level, declarative languages seem to offer a natural way of capturing human intuition. Transferring knowledge between high-level languages and low-level models is a key area of investigation for future symbiotic systems and a fruitful domain for combining data-driven and symbolic approaches. 

\subsection{Understanding Causality}

Still, the most demanding outcomes that are expected by the integration of knowledge-based and ML methodologies concern the aspects of causality learning and explainability. Existing works on harvesting causality knowledge do not yet offer convincing models. As argued in~\cite{Pearl18}, ML needs to go beyond the detection of associations, in order to exhibit explainability and counterfactual reasoning. 

The black-box character of ML-based methods hinders the understanding of their behavior, and eventually the acceptance of such systems. For example, recent studies demonstrate the fundamental inability of neural networks to efficiently and robustly learn visual relations, which renders the high performance that networks of this type often achieve worth a closer investigation~\cite{kim2018not,rosenfeld2018elephant}. 
Advancement in exploiting CS knowledge is expected to offer a significant leverage in understanding and reasoning with causal relations. And, of course, transparent reasoning is vital in understanding the abilities and constrains of existing systems. Yet, as indicated in the current review, this latter direction is still not pursued in a coordinated and structured way.

\subsection{Achieving a Tighter Integration}

Ultimately, unifying logical and probabilistic graphical models seems to be at the heart of handling the majority of real-world problems. Recent studies show that even a loosely-coupled integration can achieve better accuracy in classification problems with small datasets in comparison with end-to-end deep networks and comparable accuracy with larger datasets (see e.g., \cite{RileySridharan19,BasuShakerinGupta20}). A tighter integration is highly anticipated, as it will help build systems that learn from data, while still being able to generalize to domains other than the ones trained for. Existing solutions are indeed promising, as for example approaches based on the widely used Markov Logic, which nevertheless introduces limitations on both the theoretical and the practical level \cite{DomingosLowd19}. Its first-order nature, for instance, often contradicts with the non-monotonicity met in CS domains. The support for complex tasks, such as causal, temporal or counterfactual reasoning, in a non-monotonic fashion and over rich conceptual representations unfolds a series of research questions worth exploring in the near future.

%% file: conclusions.tex

\section{Conclusions}

In this paper, we reviewed approaches that rely on both knowledge-based and data-driven methods, in order to offer solutions to the field of intelligent object perception. By adopting a knowledge-driven, rather than a problem-specific grouping, we analyzed a multitude of approaches that attempt to unify high-level knowledge with diverse machine learning systems. The review revealed open and prominent directions, showing clear evidence that hybrid methods constitute an avenue worth exploring.


%% file: aaaimake20Survey.bbl
\begin{thebibliography}{}

\bibitem[\protect\citeauthoryear{Aditya \bgroup et al\mbox.\egroup
  }{2018}]{Aditya2018}
Aditya, S.; Yang, Y.; Baral, C.; Aloimonos, Y.; and Ferm{\"{u}}ller, C.
\newblock 2018.
\newblock {Image Understanding using vision and reasoning through Scene
  Description Graph}.
\newblock {\em Computer Vision and Image Understanding} 173:33--45.

\bibitem[\protect\citeauthoryear{Aditya, Yang, and Baral}{2019}]{Somak19}
Aditya, S.; Yang, Y.; and Baral, C.
\newblock 2019.
\newblock Integrating knowledge and reasoning in image understanding.
\newblock In {\em Proceedings of the Twenty-Eighth International Joint
  Conference on Artificial Intelligence, {IJCAI-19}},  6252--6259.
\newblock International Joint Conferences on Artificial Intelligence
  Organization.

\bibitem[\protect\citeauthoryear{Agostini, Torras, and
  Woergoetter}{2017}]{agostini2017efficient}
Agostini, A.; Torras, C.; and Woergoetter, F.
\newblock 2017.
\newblock Efficient interactive decision-making framework for robotic
  applications.
\newblock {\em Artificial Intelligence} 247:187--212.

\bibitem[\protect\citeauthoryear{Antanas \bgroup et al\mbox.\egroup
  }{2018a}]{Antanas18}
Antanas, L.; Dries, A.; Moreno, P.; and De~Raedt, L.
\newblock 2018a.
\newblock Relational affordance learning for task-dependent robot grasping.
\newblock In Lachiche, N., and Vrain, C., eds., {\em Inductive Logic
  Programming},  1--15.
\newblock Cham: Springer International Publishing.

\bibitem[\protect\citeauthoryear{Antanas \bgroup et al\mbox.\egroup
  }{2018b}]{antanas2018semantic}
Antanas, L.; Moreno, P.; Neumann, M.; de~Figueiredo, R.~P.; Kersting, K.;
  Santos-Victor, J.; and De~Raedt, L.
\newblock 2018b.
\newblock Semantic and geometric reasoning for robotic grasping: a
  probabilistic logic approach.
\newblock {\em Autonomous Robots}  1--26.

\bibitem[\protect\citeauthoryear{Basu, Shakerin, and
  Gupta}{2020}]{BasuShakerinGupta20}
Basu, K.; Shakerin, F.; and Gupta, G.
\newblock 2020.
\newblock Aqua: Asp-based visual question answering.
\newblock In Komendantskaya, E., and Liu, Y.~A., eds., {\em Practical Aspects
  of Declarative Languages},  57--72.
\newblock Springer International Publishing.

\bibitem[\protect\citeauthoryear{Beetz \bgroup et al\mbox.\egroup
  }{2018}]{beetz2018know}
Beetz, M.; Be{\ss}ler, D.; Haidu, A.; Pomarlan, M.; Bozcuo{\u{g}}lu, A.~K.; and
  Bartels, G.
\newblock 2018.
\newblock Know rob 2.0âa 2nd generation knowledge processing framework for
  cognition-enabled robotic agents.
\newblock In {\em 2018 IEEE ICRA},  512--519.
\newblock IEEE.

\bibitem[\protect\citeauthoryear{Bengio \bgroup et al\mbox.\egroup
  }{2019}]{Bengio19}
Bengio, Y.; Deleu, T.; Rahaman, N.; Ke, R.; Lachapelle, S.; Bilaniuk, O.;
  Goyal, A.; and Pal, C.
\newblock 2019.
\newblock A meta-transfer objective for learning to disentangle causal
  mechanisms.
\newblock {\em arXiv preprint arXiv:1901.10912}.

\bibitem[\protect\citeauthoryear{Brachman and Levesque}{2004}]{Brachman04book}
Brachman, R., and Levesque, H.
\newblock 2004.
\newblock {\em Knowledge Representation and Reasoning}.
\newblock San Francisco, CA, USA: Morgan Kaufmann Publishers Inc.

\bibitem[\protect\citeauthoryear{Chao \bgroup et al\mbox.\egroup
  }{2015}]{Chao2015}
Chao, Y.~W.; Wang, Z.; He, Y.; Wang, J.; and Deng, J.
\newblock 2015.
\newblock {HICO: A benchmark for recognizing human-object interactions in
  images}.
\newblock {\em IEEE ICCV} 2015 Inter:1017--1025.

\bibitem[\protect\citeauthoryear{Chen \bgroup et al\mbox.\egroup
  }{2018}]{Chen2018}
Chen, X.; Li, L.~J.; Fei-Fei, L.; and Gupta, A.
\newblock 2018.
\newblock {Iterative Visual Reasoning beyond Convolutions}.
\newblock {\em IEEE CVPR}  7239--7248.

\bibitem[\protect\citeauthoryear{Chernova \bgroup et al\mbox.\egroup
  }{2017}]{chernova2017situated}
Chernova, S.; Chu, V.; Daruna, A.; Garrison, H.; Hahn, M.; Khante, P.; Liu, W.;
  and Thomaz, A.
\newblock 2017.
\newblock Situated bayesian reasoning framework for robots operating in diverse
  everyday environments.
\newblock In {\em International Symposium on Robotics Research (ISRR)}.

\bibitem[\protect\citeauthoryear{Chuang \bgroup et al\mbox.\egroup
  }{2018}]{Chuang2018}
Chuang, C.~Y.; Li, J.; Torralba, A.; and Fidler, S.
\newblock 2018.
\newblock {Learning to Act Properly: Predicting and Explaining Affordances from
  Images}.
\newblock {\em IEEE CVPR}  975--983.

\bibitem[\protect\citeauthoryear{Daruna \bgroup et al\mbox.\egroup
  }{2019}]{daruna2019robocse}
Daruna, A.; Liu, W.; Kira, Z.; and Chernova, S.
\newblock 2019.
\newblock Robocse: Robot common sense embedding.
\newblock {\em arXiv preprint arXiv:1903.00412}.

\bibitem[\protect\citeauthoryear{Davis and Marcus}{2015}]{Davis15}
Davis, E., and Marcus, G.
\newblock 2015.
\newblock Commonsense reasoning and commonsense knowledge in artificial
  intelligence.
\newblock {\em Commun. ACM} 58(9):92--103.

\bibitem[\protect\citeauthoryear{Dean, Allen, and Aloimonos}{1995}]{Dean95book}
Dean, T.; Allen, J.; and Aloimonos, Y.
\newblock 1995.
\newblock {\em Artificial Intelligence: Theory and Practice}.
\newblock Redwood City, CA, USA: Benjamin-Cummings Publishing Co., Inc.

\bibitem[\protect\citeauthoryear{Deng \bgroup et al\mbox.\egroup
  }{2014}]{deng2014large}
Deng, J.; Ding, N.; Jia, Y.; Frome, A.; Murphy, K.; Bengio, S.; Li, Y.; Neven,
  H.; and Adam, H.
\newblock 2014.
\newblock Large-scale object classification using label relation graphs.
\newblock In {\em ECCV},  48--64.
\newblock Springer.

\bibitem[\protect\citeauthoryear{Domingos and Lowd}{2019}]{DomingosLowd19}
Domingos, P., and Lowd, D.
\newblock 2019.
\newblock Unifying logical and statistical ai with markov logic.
\newblock {\em Commununications of the ACM} 62(7):74–83.

\bibitem[\protect\citeauthoryear{Fu \bgroup et al\mbox.\egroup
  }{2015}]{fu2015zero}
Fu, Z.; Xiang, T.; Kodirov, E.; and Gong, S.
\newblock 2015.
\newblock Zero-shot object recognition by semantic manifold distance.
\newblock In {\em IEEE CVPR},  2635--2644.

\bibitem[\protect\citeauthoryear{Geffner}{2018}]{Geffner18}
Geffner, H.
\newblock 2018.
\newblock Model-free, model-based, and general intelligence.
\newblock In {\em Proceedings of the 27th International Joint Conference on
  Artificial Intelligence}, IJCAI'18,  10--17.
\newblock AAAI Press.

\bibitem[\protect\citeauthoryear{Gemignani \bgroup et al\mbox.\egroup
  }{2016}]{gemignani2016living}
Gemignani, G.; Capobianco, R.; Bastianelli, E.; Bloisi, D.~D.; Iocchi, L.; and
  Nardi, D.
\newblock 2016.
\newblock Living with robots: Interactive environmental knowledge acquisition.
\newblock {\em Robotics and Autonomous Systems} 78:1--16.

\bibitem[\protect\citeauthoryear{Goyal \bgroup et al\mbox.\egroup
  }{2019}]{Goyal2019}
Goyal, Y.; Khot, T.; Agrawal, A.; Summers-Stay, D.; Batra, D.; and Parikh, D.
\newblock 2019.
\newblock {Making the V in VQA Matter: Elevating the Role of Image
  Understanding in Visual Question Answering}.
\newblock {\em IJCV} 127(4):398--414.

\bibitem[\protect\citeauthoryear{Gu \bgroup et al\mbox.\egroup
  }{2019}]{gu2019scene}
Gu, J.; Zhao, H.; Lin, Z.; Li, S.; Cai, J.; and Ling, M.
\newblock 2019.
\newblock Scene graph generation with external knowledge and image
  reconstruction.
\newblock In {\em IEEE CVPR},  1969--1978.

\bibitem[\protect\citeauthoryear{Herath, Harandi, and
  Porikli}{2017}]{herath2017going}
Herath, S.; Harandi, M.; and Porikli, F.
\newblock 2017.
\newblock Going deeper into action recognition: A survey.
\newblock {\em IMAVIS} 60:4--21.

\bibitem[\protect\citeauthoryear{Hu \bgroup et al\mbox.\egroup }{2017}]{Hu2017}
Hu, R.; Andreas, J.; Rohrbach, M.; Darrell, T.; and Saenko, K.
\newblock 2017.
\newblock {Learning to Reason: End-to-End Module Networks for Visual Question
  Answering}.
\newblock {\em IEEE ICCV} 2017-Octob(Figure 1):804--813.

\bibitem[\protect\citeauthoryear{Icarte \bgroup et al\mbox.\egroup
  }{2017}]{icarte2017general}
Icarte, R.~T.; Baier, J.~A.; Ruz, C.; and Soto, A.
\newblock 2017.
\newblock How a general-purpose commonsense ontology can improve performance of
  learning-based image retrieval.
\newblock {\em arXiv preprint arXiv:1705.08844}.

\bibitem[\protect\citeauthoryear{Johnson \bgroup et al\mbox.\egroup
  }{2017}]{johnson2017inferring}
Johnson, J.; Hariharan, B.; van~der Maaten, L.; Hoffman, J.; Fei-Fei, L.;
  Lawrence~Zitnick, C.; and Girshick, R.
\newblock 2017.
\newblock Inferring and executing programs for visual reasoning.
\newblock In {\em IEEE ICCV},  2989--2998.

\bibitem[\protect\citeauthoryear{Katzouris \bgroup et al\mbox.\egroup
  }{2019}]{Katzouris19}
Katzouris, N.; Michelioudakis, E.; Artikis, A.; and Paliouras, G.
\newblock 2019.
\newblock Online learning of weighted relational rules for complex event
  recognition.
\newblock In Berlingerio, M.; Bonchi, F.; G{\"a}rtner, T.; Hurley, N.; and
  Ifrim, G., eds., {\em Machine Learning and Knowledge Discovery in Databases},
   396--413.
\newblock Cham: Springer International Publishing.

\bibitem[\protect\citeauthoryear{Kim, Ricci, and Serre}{2018}]{kim2018not}
Kim, J.; Ricci, M.; and Serre, T.
\newblock 2018.
\newblock Not-so-clevr: learning same--different relations strains feedforward
  neural networks.
\newblock {\em Interface focus} 8(4):20180011.

\bibitem[\protect\citeauthoryear{Krishna \bgroup et al\mbox.\egroup
  }{2017}]{Krishna2017}
Krishna, R.; Zhu, Y.; Groth, O.; Johnson, J.; Hata, K.; Kravitz, J.; Chen, S.;
  Kalantidis, Y.; Li, L.~J.; Shamma, D.~A.; Bernstein, M.~S.; and Fei-Fei, L.
\newblock 2017.
\newblock {Visual Genome: Connecting Language and Vision Using Crowdsourced
  Dense Image Annotations}.
\newblock {\em IJCV} 123(1):32--73.

\bibitem[\protect\citeauthoryear{Lee \bgroup et al\mbox.\egroup
  }{2018}]{Lee2018}
Lee, C.~W.; Fang, W.; Yeh, C.~K.; and Wang, Y. C.~F.
\newblock 2018.
\newblock {Multi-label Zero-Shot Learning with Structured Knowledge Graphs}.
\newblock {\em IEEE CVPR}  1576--1585.

\bibitem[\protect\citeauthoryear{Lemaignan \bgroup et al\mbox.\egroup
  }{2017}]{lemaignan2017artificial}
Lemaignan, S.; Warnier, M.; Sisbot, E.~A.; Clodic, A.; and Alami, R.
\newblock 2017.
\newblock Artificial cognition for social human--robot interaction: An
  implementation.
\newblock {\em Artificial Intelligence} 247:45--69.

\bibitem[\protect\citeauthoryear{Li \bgroup et al\mbox.\egroup
  }{2015}]{li2015gated}
Li, Y.; Tarlow, D.; Brockschmidt, M.; and Zemel, R.
\newblock 2015.
\newblock Gated graph sequence neural networks.
\newblock {\em arXiv preprint arXiv:1511.05493}.

\bibitem[\protect\citeauthoryear{Li \bgroup et al\mbox.\egroup
  }{2017}]{li2017situation}
Li, R.; Tapaswi, M.; Liao, R.; Jia, J.; Urtasun, R.; and Fidler, S.
\newblock 2017.
\newblock Situation recognition with graph neural networks.
\newblock In {\em IEEE ICCV},  4173--4182.

\bibitem[\protect\citeauthoryear{Li \bgroup et al\mbox.\egroup
  }{2019}]{li2019visual}
Li, H.; Wang, P.; Shen, C.; and Hengel, A. v.~d.
\newblock 2019.
\newblock Visual question answering as reading comprehension.
\newblock In {\em IEEE CVPR},  6319--6328.

\bibitem[\protect\citeauthoryear{Li, Su, and Zhu}{2017}]{li2017incorporating}
Li, G.; Su, H.; and Zhu, W.
\newblock 2017.
\newblock Incorporating external knowledge to answer open-domain visual
  questions with dynamic memory networks.
\newblock {\em arXiv preprint arXiv:1712.00733}.

\bibitem[\protect\citeauthoryear{Lin and Parikh}{2015}]{lin2015don}
Lin, X., and Parikh, D.
\newblock 2015.
\newblock Don't just listen, use your imagination: Leveraging visual common
  sense for non-visual tasks.
\newblock In {\em IEEE CVPR},  2984--2993.

\bibitem[\protect\citeauthoryear{Liu \bgroup et al\mbox.\egroup
  }{2018a}]{liu2018deep}
Liu, L.; Ouyang, W.; Wang, X.; Fieguth, P.; Chen, J.; Liu, X.; and
  Pietik{\"a}inen, M.
\newblock 2018a.
\newblock Deep learning for generic object detection: A survey.
\newblock {\em arXiv preprint arXiv:1809.02165}.

\bibitem[\protect\citeauthoryear{Liu \bgroup et al\mbox.\egroup
  }{2018b}]{Liu2018}
Liu, Y.; Wang, R.; Shan, S.; and Chen, X.
\newblock 2018b.
\newblock {Structure Inference Net: Object Detection Using Scene-Level Context
  and Instance-Level Relationships}.
\newblock {\em IEEE CVPR}  6985--6994.

\bibitem[\protect\citeauthoryear{Lu \bgroup et al\mbox.\egroup }{2016}]{Lu2016}
Lu, C.; Krishna, R.; Bernstein, M.; and Fei-Fei, L.
\newblock 2016.
\newblock {Visual relationship detection with language priors}.
\newblock {\em Lecture Notes in Computer Science (including subseries Lecture
  Notes in Artificial Intelligence and Lecture Notes in Bioinformatics)} 9905
  LNCS(Figure 2):852--869.

\bibitem[\protect\citeauthoryear{Lu \bgroup et al\mbox.\egroup }{2018}]{Lu2018}
Lu, P.; Ji, L.; Zhang, W.; Duan, N.; Zhou, M.; and Wang, J.
\newblock 2018.
\newblock {R-VQA: Learning visual relation facts with semantic attention for
  visual question answering}.
\newblock {\em Proceedings of the ACM SIGKDD International Conference on
  Knowledge Discovery and Data Mining}  1880--1889.

\bibitem[\protect\citeauthoryear{Marino, Salakhutdinov, and
  Gupta}{2017}]{Marino2017}
Marino, K.; Salakhutdinov, R.; and Gupta, A.
\newblock 2017.
\newblock {The more you know: using knowledge graphs for image classification}.
\newblock {\em IEEE CVPR} 2017-Janua:20--28.

\bibitem[\protect\citeauthoryear{Moldovan \bgroup et al\mbox.\egroup
  }{2018}]{Moldovan2018}
Moldovan, B.; Moreno, P.; Nitti, D.; Santos-Victor, J.; and De~Raedt, L.
\newblock 2018.
\newblock Relational affordances for multiple-object manipulation.
\newblock {\em Autonomous Robots} 42(1):19--44.

\bibitem[\protect\citeauthoryear{Narasimhan and
  Schwing}{2018}]{narasimhan2018straight}
Narasimhan, M., and Schwing, A.~G.
\newblock 2018.
\newblock Straight to the facts: Learning knowledge base retrieval for factual
  visual question answering.
\newblock In {\em Proceedings of the ECCV (ECCV)},  451--468.

\bibitem[\protect\citeauthoryear{Nickel \bgroup et al\mbox.\egroup
  }{2016}]{Nickel0TG16}
Nickel, M.; Murphy, K.; Tresp, V.; and Gabrilovich, E.
\newblock 2016.
\newblock A review of relational machine learning for knowledge graphs.
\newblock {\em Proceedings of the {IEEE}} 104(1):11--33.

\bibitem[\protect\citeauthoryear{Pearl}{2018}]{Pearl18}
Pearl, J.
\newblock 2018.
\newblock Theoretical impediments to machine learning with seven sparks from
  the causal revolution.
\newblock {\em arXiv preprint arXiv:1801.04016}.

\bibitem[\protect\citeauthoryear{Rajan and Saffiotti}{2017}]{AIRoboticsSP17}
Rajan, K., and Saffiotti, A., eds.
\newblock 2017.
\newblock {\em Special Issue on AI and Robotics}, volume 247.
\newblock Elsevier.
\newblock  1--440.

\bibitem[\protect\citeauthoryear{Ramanathan \bgroup et al\mbox.\egroup
  }{2015}]{Ramanathan2015}
Ramanathan, V.; Li, C.; Deng, J.; and Han, W.
\newblock 2015.
\newblock {Learning semantic relationships for better action retrieval in
  images ( Supplementary )}.
\newblock {\em Computer Vision and Pattern Recognition}  1--4.

\bibitem[\protect\citeauthoryear{Ramirez-Amaro, Beetz, and
  Cheng}{2017}]{ramirez2017transferring}
Ramirez-Amaro, K.; Beetz, M.; and Cheng, G.
\newblock 2017.
\newblock Transferring skills to humanoid robots by extracting semantic
  representations from observations of human activities.
\newblock {\em Artificial Intelligence} 247:95--118.

\bibitem[\protect\citeauthoryear{Ravichandar \bgroup et al\mbox.\egroup
  }{2019}]{ravichandar2019lfd}
Ravichandar, H.; Polydoros, A.~S.; Chernova, S.; and Billard, A.
\newblock 2019.
\newblock Robot learning from demonstration: A review of recent advances.
\newblock {\em Annual Review of Control, Robotics, and Autonomous Systems}  In
  Press.

\bibitem[\protect\citeauthoryear{Redmon and Farhadi}{2017}]{redmon2017yolo9000}
Redmon, J., and Farhadi, A.
\newblock 2017.
\newblock Yolo9000: better, faster, stronger.
\newblock In {\em IEEE CVPR},  7263--7271.

\bibitem[\protect\citeauthoryear{Riley and Sridharan}{2019}]{RileySridharan19}
Riley, H., and Sridharan, M.
\newblock 2019.
\newblock Integrating non-monotonic logical reasoning and inductive learning
  with deep learning for explainable visual question answering.
\newblock {\em Frontiers in Robotics and AI} 6:125.

\bibitem[\protect\citeauthoryear{Rosenfeld, Zemel, and
  Tsotsos}{2018}]{rosenfeld2018elephant}
Rosenfeld, A.; Zemel, R.; and Tsotsos, J.~K.
\newblock 2018.
\newblock The elephant in the room.
\newblock {\em arXiv preprint arXiv:1808.03305}.

\bibitem[\protect\citeauthoryear{Ruiz-Sarmiento, Galindo, and
  Gonzalez-Jimenez}{2016}]{ruiz2016probability}
Ruiz-Sarmiento, J.-R.; Galindo, C.; and Gonzalez-Jimenez, J.
\newblock 2016.
\newblock Probability and common-sense: Tandem towards robust robotic object
  recognition in ambient assisted living.
\newblock In {\em Ubiquitous Computing and Ambient Intelligence}. Springer.
\newblock  3--8.

\bibitem[\protect\citeauthoryear{Sadeghi, Kumar~Divvala, and
  Farhadi}{2015}]{sadeghi2015viske}
Sadeghi, F.; Kumar~Divvala, S.~K.; and Farhadi, A.
\newblock 2015.
\newblock Viske: Visual knowledge extraction and question answering by visual
  verification of relation phrases.
\newblock In {\em IEEE CVPR},  1456--1464.

\bibitem[\protect\citeauthoryear{Santoro \bgroup et al\mbox.\egroup
  }{2017}]{Santoro2017}
Santoro, A.; Raposo, D.; Barrett, D. G.~T.; Malinowski, M.; Pascanu, R.;
  Battaglia, P.; and Lillicrap, T.
\newblock 2017.
\newblock {A simple neural network module for relational reasoning}.
\newblock (Nips).

\bibitem[\protect\citeauthoryear{Sawatzky \bgroup et al\mbox.\egroup
  }{2019}]{Sawatzky2019}
Sawatzky, J.; Souri, Y.; Grund, C.; and Gall, J.
\newblock 2019.
\newblock {What Object Should I Use? - Task Driven Object Detection}.

\bibitem[\protect\citeauthoryear{Scarselli \bgroup et al\mbox.\egroup
  }{2008}]{scarselli2008graph}
Scarselli, F.; Gori, M.; Tsoi, A.~C.; Hagenbuchner, M.; and Monfardini, G.
\newblock 2008.
\newblock The graph neural network model.
\newblock {\em IEEE Transactions on NN} 20(1):61--80.

\bibitem[\protect\citeauthoryear{Shah \bgroup et al\mbox.\egroup
  }{2019}]{shah2019kvqa}
Shah, S.; Mishra, A.; Yadati, N.; and Talukdar, P.~P.
\newblock 2019.
\newblock Kvqa: Knowledge-aware visual question answering.
\newblock AAAI.

\bibitem[\protect\citeauthoryear{Stone \bgroup et al\mbox.\egroup
  }{2016}]{Stone16AIReport}
Stone, P.; Brooks, R.; Brynjolfsson, E.; Calo, R.; Etzioni, O.; Hager, G.;
  Hirschberg, J.; Kalyanakrishnan, S.; Kamar, E.; Kraus, S.; Leyton-Brown, K.;
  Parkes, D.; Press, W.; Saxenian, A.; Shah, J.; Tambe, M.; ; and Teller, A.
\newblock 2016.
\newblock Artificial intelligence and life in 2030.
\newblock {\em One Hundred Year Study on Artificial Intelligence: Report of the
  2015-2016 Study Panel}.

\bibitem[\protect\citeauthoryear{Su \bgroup et al\mbox.\egroup }{2018}]{Su2018}
Su, Z.; Zhu, C.; Dong, Y.; Cai, D.; Chen, Y.; and Li, J.
\newblock 2018.
\newblock {Learning Visual Knowledge Memory Networks for Visual Question
  Answering}.
\newblock {\em IEEE CVPR}  7736--7745.

\bibitem[\protect\citeauthoryear{Suchan \bgroup et al\mbox.\egroup
  }{2017}]{SuchanMehul17}
Suchan, J.; Bhatt, M.; Walega, P.~A.; and Schultz, C. P.~L.
\newblock 2017.
\newblock Visual explanation by high-level abduction: On answer-set programming
  driven reasoning about moving objects.
\newblock {\em CoRR} abs/1712.00840.

\bibitem[\protect\citeauthoryear{Tenorth and
  Beetz}{2017}]{tenorth2017representations}
Tenorth, M., and Beetz, M.
\newblock 2017.
\newblock Representations for robot knowledge in the knowrob framework.
\newblock {\em Artificial Intelligence} 247:151--169.

\bibitem[\protect\citeauthoryear{Torabi, Warnell, and Stone}{2019}]{Torabi19}
Torabi, F.; Warnell, G.; and Stone, P.
\newblock 2019.
\newblock Recent advances in imitation learning from observation.
\newblock In {\em Proceedings of the Twenty-Eighth International Joint
  Conference on Artificial Intelligence, {IJCAI-19}},  6325--6331.
\newblock International Joint Conferences on Artificial Intelligence
  Organization.

\bibitem[\protect\citeauthoryear{van Harmelen \bgroup et al\mbox.\egroup
  }{2007}]{vanHarmelenLifHandbook07}
van Harmelen, F.; van Harmelen, F.; Lifschitz, V.; and Porter, B.
\newblock 2007.
\newblock {\em Handbook of Knowledge Representation}.
\newblock San Diego, USA: Elsevier Science.

\bibitem[\protect\citeauthoryear{Vedantam \bgroup et al\mbox.\egroup
  }{2015}]{vedantam2015learning}
Vedantam, R.; Lin, X.; Batra, T.; Lawrence~Zitnick, C.; and Parikh, D.
\newblock 2015.
\newblock Learning common sense through visual abstraction.
\newblock In {\em IEEE ICCV},  2542--2550.

\bibitem[\protect\citeauthoryear{Wang \bgroup et al\mbox.\egroup
  }{2018}]{wang2018fvqa}
Wang, P.; Wu, Q.; Shen, C.; Dick, A.; and van~den Hengel, A.
\newblock 2018.
\newblock Fvqa: Fact-based visual question answering.
\newblock {\em IEEE Trans. on PAMI} 40(10):2413--2427.

\bibitem[\protect\citeauthoryear{Wang, Ye, and Gupta}{2018}]{Wang2018b}
Wang, X.; Ye, Y.; and Gupta, A.
\newblock 2018.
\newblock {Zero-Shot Recognition via Semantic Embeddings and Knowledge Graphs}.
\newblock {\em IEEE CVPR}  6857--6866.

\bibitem[\protect\citeauthoryear{Wu \bgroup et al\mbox.\egroup
  }{2016a}]{wu2016ask}
Wu, Q.; Wang, P.; Shen, C.; Dick, A.; and van~den Hengel, A.
\newblock 2016a.
\newblock Ask me anything: Free-form visual question answering based on
  knowledge from external sources.
\newblock In {\em IEEE CVPR},  4622--4630.

\bibitem[\protect\citeauthoryear{Wu \bgroup et al\mbox.\egroup
  }{2016b}]{Wu_2016_CVPR}
Wu, Z.; Fu, Y.; Jiang, Y.-G.; and Sigal, L.
\newblock 2016b.
\newblock Harnessing object and scene semantics for large-scale video
  understanding.
\newblock In {\em TheIEEE CVPR}.

\bibitem[\protect\citeauthoryear{Wu \bgroup et al\mbox.\egroup
  }{2017}]{wu2017visual}
Wu, Q.; Teney, D.; Wang, P.; Shen, C.; Dick, A.; and van~den Hengel, A.
\newblock 2017.
\newblock Visual question answering: A survey of methods and datasets.
\newblock {\em CVIU} 163:21--40.

\bibitem[\protect\citeauthoryear{Wu \bgroup et al\mbox.\egroup }{2018}]{Wu2018}
Wu, Q.; Shen, C.; Wang, P.; Dick, A.; and {Van Den Hengel}, A.
\newblock 2018.
\newblock {Image Captioning and Visual Question Answering Based on Attributes
  and External Knowledge}.
\newblock {\em IEEE Trans. on PAMI} 40(6):1367--1381.

\bibitem[\protect\citeauthoryear{Xian \bgroup et al\mbox.\egroup
  }{2016}]{xian2016latent}
Xian, Y.; Akata, Z.; Sharma, G.; Nguyen, Q.; Hein, M.; and Schiele, B.
\newblock 2016.
\newblock Latent embeddings for zero-shot classification.
\newblock In {\em IEEE CVPR},  69--77.

\bibitem[\protect\citeauthoryear{Ye \bgroup et al\mbox.\egroup }{2017}]{Ye2017}
Ye, C.; Yang, Y.; Mao, R.; Fermuller, C.; and Aloimonos, Y.
\newblock 2017.
\newblock {What can i do around here? Deep functional scene understanding for
  cognitive robots}.
\newblock {\em IEEE ICRA}  4604--4611.

\bibitem[\protect\citeauthoryear{Yi \bgroup et al\mbox.\egroup }{2018}]{Yi2018}
Yi, K.; Torralba, A.; Wu, J.; Kohli, P.; Gan, C.; and Tenenbaum, J.~B.
\newblock 2018.
\newblock {Neural-symbolic VQA: Disentangling reasoning from vision and
  language understanding}.
\newblock {\em Advances in Neural Information Processing Systems}
  2018-December(NeurIPS):1031--1042.

\bibitem[\protect\citeauthoryear{Young \bgroup et al\mbox.\egroup
  }{2016}]{young2016towards}
Young, J.; Basile, V.; Kunze, L.; Cabrio, E.; and Hawes, N.
\newblock 2016.
\newblock Towards lifelong object learning by integrating situated robot
  perception and semantic web mining.
\newblock In {\em Proceedings of the Twenty-second European Conference on
  Artificial Intelligence},  1458--1466.
\newblock IOS Press.

\bibitem[\protect\citeauthoryear{Young \bgroup et al\mbox.\egroup
  }{2017}]{young2017making}
Young, J.; Basile, V.; Suchi, M.; Kunze, L.; Hawes, N.; Vincze, M.; and Caputo,
  B.
\newblock 2017.
\newblock Making sense of indoor spaces using semantic web mining and situated
  robot perception.
\newblock In {\em European Semantic Web Conference},  299--313.
\newblock Springer.

\bibitem[\protect\citeauthoryear{Zhu \bgroup et al\mbox.\egroup
  }{2015a}]{Zhu2015a}
Zhu, Y.; Groth, O.; Bernstein, M.; and Fei-Fei, L.
\newblock 2015a.
\newblock {Visual7W: Grounded Question Answering in Images}.

\bibitem[\protect\citeauthoryear{Zhu \bgroup et al\mbox.\egroup
  }{2015b}]{zhu2015building}
Zhu, Y.; Zhang, C.; R{\'e}, C.; and Fei-Fei, L.
\newblock 2015b.
\newblock Building a large-scale multimodal knowledge base for visual question
  answering.
\newblock {\em CoRR} abs/1507.05670.

\bibitem[\protect\citeauthoryear{Zhu, Fathi, and Fei-Fei}{2014}]{Feifei14}
Zhu, Y.; Fathi, A.; and Fei-Fei, L.
\newblock 2014.
\newblock Reasoning about object affordances in a knowledge base
  representation.
\newblock In Fleet, D.; Pajdla, T.; Schiele, B.; and Tuytelaars, T., eds., {\em
  ECCV},  408--424.
\newblock Cham: Springer International Publishing.

\end{thebibliography}
